\documentclass{article} 
\usepackage{nips15submit_e,times}
\usepackage{hyperref}
\usepackage{url}
\usepackage{graphicx}
\usepackage{subcaption}

\title{Scheduled Sampling for Sequence Prediction with Recurrent Neural Networks}

\author{
Samy Bengio,
Oriol Vinyals,
Navdeep Jaitly,
Noam Shazeer \\
Google Research \\
Mountain View, CA, USA\\
\texttt{\{bengio,vinyals,ndjaitly,noam\}@google.com} \\
}

\nipsfinalcopy 

\begin{document}

\maketitle

\begin{abstract}
Recurrent Neural Networks can be trained to produce sequences of tokens given
some input, as exemplified by recent results in machine translation and image
captioning. The current approach to training them consists of maximizing the
likelihood of each token in the sequence given the current (recurrent) state
and the previous token. At inference, the unknown previous token is then
replaced by a token generated by the model itself. This discrepancy between
training and inference can yield errors that can accumulate quickly along the
generated sequence.
We propose a curriculum learning strategy to gently change the
training process from a fully guided scheme using the true previous token,
towards a less guided scheme which mostly uses the generated token instead.
Experiments on several sequence prediction tasks show that this approach
yields significant improvements. Moreover, it was used successfully
in our winning entry to the MSCOCO image captioning challenge, 2015.

\end{abstract}

\section{Introduction}
\label{sec:introduction}

Recurrent neural networks can be used to process sequences, either as input,
output or both. While they are known to be hard to train when there are long
term dependencies in the data~\cite{bengio:1994:ieee}, some versions like the
Long Short-Term Memory (LSTM)~\cite{hochreiter1997long}
are better suited for this.
In fact, they have recently shown impressive performance in
several sequence prediction problems including machine
translation~\cite{sutskever:2014}, contextual
parsing~\cite{vinyals:2014:arxiv}, image captioning~\cite{vinyals:2015}
and even video description~\cite{donahue:2015}.

In this paper, we consider the set of problems that attempt to generate
a sequence of tokens of variable size, such as the problem of machine
translation, where the goal is to translate
a given sentence from a source language to a target language. We also consider
problems in which the input is not necessarily a sequence, like the image captioning
problem, where the goal is to generate a textual description of a given image.

In both cases, recurrent neural networks (or their variants like LSTMs)
are generally trained to maximize
the likelihood of generating the target sequence of tokens given the input.
In practice, this is done by maximizing the likelihood of each target token
given the current state of the model (which summarizes the input and
the past output tokens) and the previous target token, which
helps the model learn a kind of language model over target tokens.
However, during
inference, true {\em previous} target tokens are unavailable, and are
thus replaced by tokens generated by the model itself, yielding a discrepancy
between how the model is used at training and inference. This discrepancy
can be mitigated by the use of a beam search heuristic maintaining
several generated target sequences, but for continuous state
space models like recurrent neural networks, there is no dynamic programming
approach, so the effective number of sequences considered remains small, even with
beam search.

The main problem is that mistakes made early in the sequence generation
process are fed as input to the model and can be quickly amplified because the
model might be in a part of the state space it has
never seen at training time.

Here, we propose a {\em curriculum learning} approach~\cite{bengio:2009:icml}
to gently bridge the gap between training and inference for sequence
prediction tasks using recurrent neural networks. We propose to change the
training process in order to gradually force the model to deal with its own
mistakes, as it would have to during inference.
Doing so, the model explores
more during training and is thus more robust to correct its own mistakes at
inference as it has learned to do so during training.
We will show experimentally that this approach yields better
performance on several sequence prediction tasks.

The paper is organized as follows: in Section~\ref{sec:approach}, we
present our proposed approach to better train sequence prediction tasks
with recurrent neural networks; this is followed by Section~\ref{sec:related}
which draws links to some related approaches. We then present some
experimental results in Section~\ref{sec:experiments} and conclude in
Section~\ref{sec:conclusion}.

\section{Proposed Approach}
\label{sec:approach}

We are considering supervised tasks where the training set
is given in terms of $N$ input/output pairs $\{X^i, Y^i\}_{i=1}^N$,
where $X^i$ is
the input and can be either static (like an image) or dynamic (like a sequence)
while the target output $Y^i$ is a sequence $y^i_1, y^i_2, \ldots, y^i_{T_i}$
of a variable number of tokens that belong to a fixed known dictionary.

\subsection{Model}
Given a single input/output pair $(X, Y)$, the log probability $P(Y|X)$ can be
computed as:
\begin{eqnarray}
\label{eq:next_step}
\log P(Y|X) & = & \log P(y_1^T|X) \nonumber\\
       & = & \sum_{t=1}^T \log P(y_t|y_1^{t-1}, X) \nonumber\\
\end{eqnarray}
where $Y$ is a sequence of length $T$ represented by tokens $y_1, y_2, \ldots, y_T$.
The latter term in the above equation is estimated by a recurrent
neural network with parameters $\theta$ by introducing a state vector,
$h_t$, that is a function
of the previous state, $h_{t-1}$, and the previous output token, $y_{t-1}$,
i.e.
\begin{equation}
\label{eq:prob}
\log P(y_t|y_1^{t-1}, X;\theta) = \log P(y_t|h_t;\theta)
\end{equation}
where $h_t$ is computed by a recurrent neural network as follows:
\begin{equation}
\label{eq:recurrence}
h_t = \left\{\begin{array}{ll}
  f(X;\theta) & \mbox{if } t = 1, \\
  f(h_{t-1}, y_{t-1};\theta) & \mbox{otherwise.}
  \end{array}\right.
\end{equation}
$P(y_t|h_t;\theta)$ is often implemented as a linear projection\footnote{
Although one could also use a multi-layered non-linear projection.} of
the state vector $h_t$ into a vector of scores, one for each token of the
output dictionary, followed by a softmax transformation to ensure the scores
are properly normalized (positive and sum to 1).
$f(h,y)$ is usually a non-linear function that combines the previous
state and the previous output in order to produce the current state.

This means that the model focuses on learning to output the next token
given the current state of the model AND the previous token.
Thus, the model represents the probability distribution of sequences in
the most general form - unlike Conditional Random Fields \cite{CRFS} and other models that
assume independence between between outputs at different time steps, given
latent variable states. The capacity of the model is only limited by the
representational capacity of the recurrent and feedforward layers. LSTMs,
with their ability to learn long range structure are especially well suited to
this task and make it possible to learn rich distributions over sequences.

In order to learn variable length sequences, a special token, $<$EOS$>$,
that signifies the end of a sequence is added to the dictionary and the model.
During training, $<$EOS$>$ is concatenated to the end of each sequence.
During inference, the model generates tokens until it generates $<$EOS$>$.

\subsection{Training}
Training recurrent neural networks to solve such tasks is usually
accomplished by using mini-batch stochastic gradient descent to look for
a set of parameters $\theta^\star$ that
maximizes the
log likelihood of producing the correct target sequence $Y^i$ given
the input data $X^i$ for all training pairs $(X^i, Y^i)$:
\begin{equation}
\label{eq:training}
\theta^\star = \arg\max_\theta \sum_{(X^i, Y^i)} \log P(Y^i|X^i;\theta)\;.
\end{equation}

\subsection{Inference}
During inference the model can generate the full sequence $y_1^T$ given $X$ by
generating one token at a time, and advancing time by one step.  When
an $<$EOS$>$ token is generated, it signifies the end of the sequence.
For this process, at time $t$, the model needs as input the output
token $y_{t-1}$ from the last time step in order to produce $y_t$. Since
we do not have access to the true previous
token, we can instead either select the most likely one given our model,
or sample according to it.

Searching for the sequence $Y$ with the highest probability given $X$
is too expensive because of the combinatorial growth in the number of sequences.
Instead we use a beam searching procedure to generate $k$ ``best'' sequences. We
do this by maintaining a heap of $m$ best candidate sequences. At each time step
new candidates are generated by extending each candidate by one token and adding
them to the heap. At the end of the step, the heap is re-pruned to only keep $m$
candidates. The beam searching is truncated when no new sequences are added, and
$k$ best sequences are returned.

While beam search is often used for discrete state based models like Hidden
Markov Models where dynamic programming can be used, it is harder to use
efficiently for continuous state based
models like recurrent neural networks, since there is no way to factor the
followed state paths in a continuous space, and hence the actual number of
candidates that can be kept during beam search decoding is very small.

In all these cases, if a wrong decision is taken at time $t-1$, the model
can be in a part of the state space that is very different from those visited from
the training distribution and for which it doesn't know what to do. Worse,
it can easily lead to cumulative bad decisions - a classic problem in sequential Gibbs
sampling type approaches to sampling, where future samples can have no influence
on the past.

\subsection{Bridging the Gap with Scheduled Sampling}

The main difference between training and inference for sequence prediction tasks
when predicting token $y_t$ is whether we use the true previous token
$y_{t-1}$ or an estimate $\hat{y}_{t-1}$ coming from the model itself.

\begin{figure}
\centering
\begin{minipage}{.6\columnwidth}
  \centering
  \includegraphics[width=0.8\linewidth]{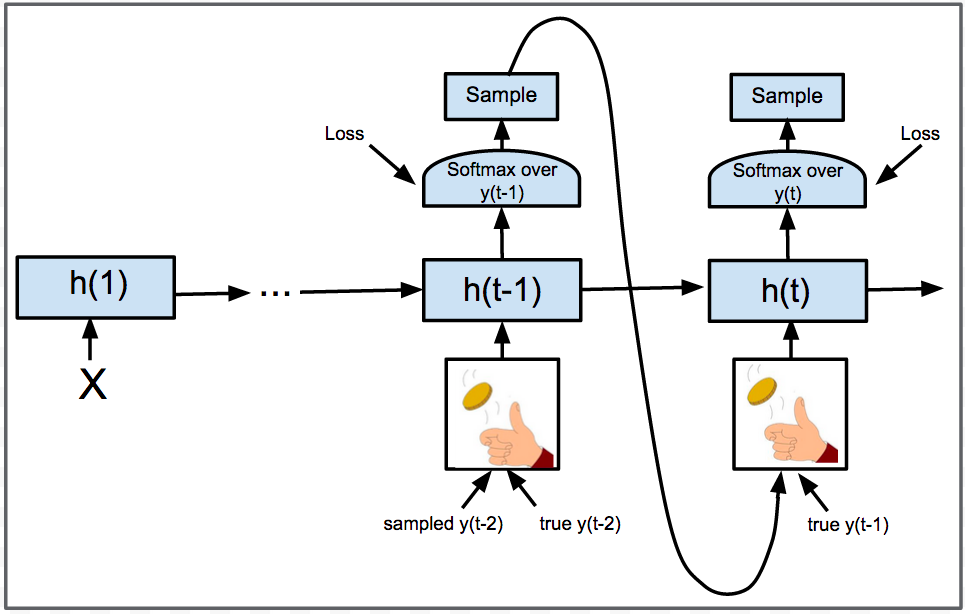}
  \captionof{figure}{\label{fig:model}Illustration of the Scheduled Sampling approach, where one flips a coin at every time step to decide to use the true previous token or one sampled from the model itself.}
\end{minipage}%
\hspace{0.3cm}
\begin{minipage}{0.35\columnwidth}
  \centering
  \includegraphics[width=\linewidth]{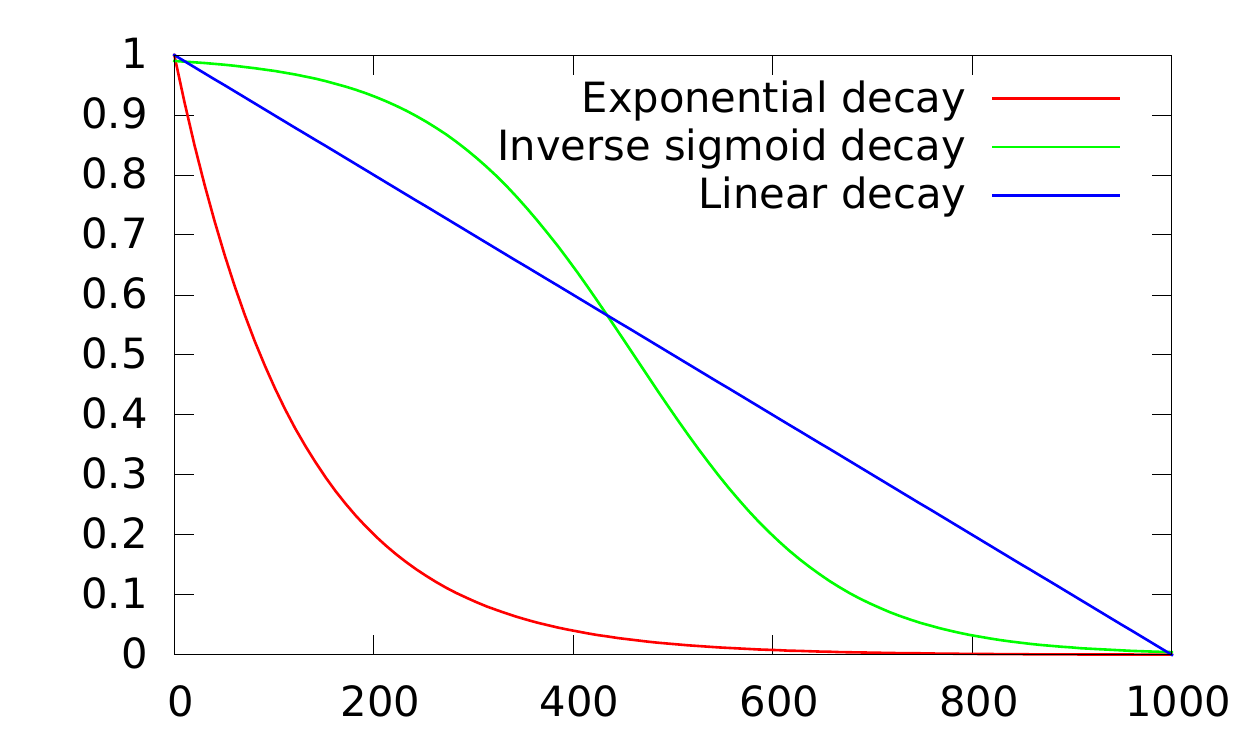}
  \captionof{figure}{\label{fig:decay}Examples of decay schedules.}
\end{minipage}
\end{figure}

We propose here a sampling mechanism that will randomly decide, during training,
whether we use $y_{t-1}$ or $\hat{y}_{t-1}$. Assuming we use a mini-batch
based stochastic gradient descent approach, for every token to predict
$y_t \in Y$ of the
$i^{th}$ mini-batch of the training algorithm, we propose to flip
a coin and use the true previous token with probability $\epsilon_i$, or an 
estimate coming from the model itself with probability
$(1 - \epsilon_i)$\footnote{Note that in the experiments, we flipped the
coin for every token. We also tried to flip the coin once per sequence, but the
results were much worse, most probably because consecutive errors are amplified during the first
rounds of training.}
The estimate of the model can be obtained by sampling a token
according to the probability distribution modeled by $P(y_{t-1}|h_{t-1})$,
or can be taken as the $\arg\max_s P(y_{t-1} = s|h_{t-1})$.
This process is illustrated in Figure~\ref{fig:model}.

When $\epsilon_i = 1$, the model is trained exactly as before, while when
$\epsilon_i = 0$ the model is trained in the same setting as inference.
We propose here a {\em curriculum learning} strategy to go from one to the
other:
intuitively, at the beginning of training, sampling from the model would
yield a random token since the model is not well trained,
which could lead to very slow convergence, so selecting more often the
true previous token should help; on the other hand, at the end of training,
$\epsilon_i$ should favor sampling from the model more often, as this
corresponds to the true inference situation, and one expects the model to
already be good enough to handle it and sample reasonable tokens.

We thus propose to use a schedule to decrease $\epsilon_i$ as a function of
$i$ itself, in a similar manner used to decrease the learning
rate in most modern stochastic gradient descent approaches.
Examples of such schedules can be seen in Figure~\ref{fig:decay} as follows:
\begin{itemize}
\item Linear decay: $\epsilon_i = \max(\epsilon, k - c i)$ where
$0 \le \epsilon < 1$ 
is the minimum amount of truth to be given to the model and $k$ and $c$
provide the offset and slope of the decay, which depend on the expected speed of
convergence.
\item Exponential decay: $\epsilon_i = k^{i}$ where $k < 1$ is a constant that
depends on the expected speed of convergence.
\item Inverse sigmoid decay: $\epsilon_i = k / (k + \exp(i / k))$ where
$k \ge 1$ depends on the expected speed of convergence.
\end{itemize}
We call our approach {\em Scheduled Sampling}.
Note that when we sample the previous token $\hat{y}_{t-1}$
from the model itself while training, we could back-propagate the gradient of
the losses at times $t \rightarrow T$ through that decision. This was not
done in the experiments described in this paper and is left for future work.

\section{Related Work}
\label{sec:related}

The discrepancy between the training and inference distributions has
already been noticed in the literature, in particular for control and
reinforcement learning tasks.

SEARN~\cite{daume:2009} was proposed to tackle problems where
supervised training examples might be different from actual test examples
when each example is made of a sequence of decisions, like acting in a
complex environment where a few mistakes of the model early in the sequential
decision process might compound and yield a very poor global performance. Their
proposed approach involves a meta-algorithm where at each meta-iteration one
trains a new model according to the current {\em policy} (essentially the
expected decisions for each situation), applies it on a test set and modifies
the next iteration policy in order to account for the previous decisions
and errors. The new policy is thus a combination of the previous one and the
actual behavior of the model.

In comparison to SEARN and related ideas~\cite{ross:2011,venkatraman:aaai:2015},
our proposed approach is completely online: a single
model is trained and the {\em policy} slowly evolves during training, instead
of a batch approach, which makes it much faster to train\footnote{In fact,
in the experiments we report in this paper, our proposed approach was not
meaningfully slower (nor faster) to train than the baseline.}
Furthermore, SEARN has been proposed in the context of reinforcement
learning, while we consider the supervised learning setting trained using
stochastic gradient descent on the overall objective.

Other approaches have considered the problem from a ranking perspective,
in particular for parsing tasks~\cite{collins:2004} where the target output
is a tree. In this case, the authors proposed to use a beam search
both during training and inference, so that both phases are
aligned. The training beam is used to find the best current estimate of the model, which is compared to the guided solution (the truth) using a ranking
loss. Unfortunately, this is not feasible when using a model like a
recurrent neural network (which is now the state-of-the-art technique in many
sequential tasks), as the state sequence cannot be factored easily (because
it is a multi-dimensional continuous state) and thus
beam search is hard to use efficiently at training time (as well as inference
time, in fact).

Finally,~\cite{goldberg:coling:2012} proposed an online algorithm for parsing problems
that adapts the targets through the use of a {\em dynamic oracle} that takes into account
the decisions of the model. The trained model is
a perceptron and is thus not state-based like a recurrent neural network, and the probability of choosing
the truth is fixed during training.

\section{Experiments}
\label{sec:experiments}

We describe in this section experiments on three different tasks, in
order to show that scheduled sampling can be helpful in different settings.
We report results on image captioning, constituency parsing and speech
recognition.

\subsection{Image Captioning}
\label{sec:coco}
Image captioning has attracted a lot of attention in the past year. The task can be formulated as a mapping of an image onto a sequence of words
describing its content in some natural language,
and most proposed approaches employ some form of recurrent network structure with simple decoding schemes~\cite{vinyals:2015, donahue:2015, baidu, toronto, karpathy:2015:cvpr}. A notable exception is the system proposed in~\cite{MSR}, which does not directly optimize the log likelihood of the caption given the image, and instead proposes a pipelined approach.

Since an image can have many valid captions, the evaluation of this task is still an open problem. Some attempts have been made to design metrics that positively correlate with human evaluation~\cite{devicider}, and a common set of tools have been published by the MSCOCO team~\cite{COCO}.

We used the MSCOCO dataset from~\cite{COCO} to train our model.
We trained on 75k images and report
results on a separate development set of 5k additional images.
Each image in the corpus has 5 different captions, so the training procedure picks one at random, creates a mini-batch of examples, and optimizes the objective function defined in~(\ref{eq:training}). The image is preprocessed by a pretrained convolutional neural network (without the last classification layer) similar to the one described in~\cite{ioffe:2015}, and the resulting image embedding is treated as if it was the first word from which the model starts generating language. 
The recurrent neural network generating words is an LSTM with one layer of 512 hidden units, and
the input words are represented by embedding vectors of size 512. The number of
words in the dictionary is 8857.
We used an inverse sigmoid decay schedule for $\epsilon_i$ for the scheduled
sampling approach.

Table~\ref{tab:coco} shows the results on various metrics on the development
set. Each of these metrics is a variant of estimating the overlap between the
obtained sequence of words and the target one. Since there were 5 target
captions per image, the best result is always chosen.
To the best of our knowledge, the baseline results are consistent
(slightly better) with the current
state-of-the-art on that task. While dropout helped in terms of log likelihood
(as expected but not shown), it had a negative impact on the real metrics. On the other hand, scheduled sampling successfully trained a model more resilient to failures due to training and inference mismatch, which likely yielded higher quality captions according to all the metrics. Ensembling models also yielded better performance, both for the baseline
and the schedule sampling approach. It is also interesting to note that
a model trained while always sampling from itself (hence in a regime similar
to inference), dubbed {\em Always Sampling} in the table, yielded very poor performance, as expected because the model has a hard time learning the task in
that case. We also trained a model with scheduled sampling, but instead of sampling from the model,
we sampled from a uniform distribution, in order to verify that it was important to build on the
current model and that the performance boost was not just a simple form of regularization. We
called this {\em Uniform Scheduled Sampling} and the results are better than the baseline, but
not as good as our proposed approach.
We also experimented with flipping the coin once per sequence instead of once
per token, but the results were as poor as the {\em Always Sampling} approach.

\begin{table}[!ht]
\caption{Various metrics (the higher the better) on the MSCOCO development set for the image captioning task.}\label{tab:coco}
\centering
\begin{tabular}{|c|c|c|c|}
 \hline
Approach vs Metric & BLEU-4 & METEOR & CIDER \\
\hline
\hline
Baseline   & 28.8  & 24.2 & 89.5     \\
Baseline with Dropout & 28.1 & 23.9 & 87.0 \\
Always Sampling & 11.2 & 15.7 & 49.7 \\
Scheduled Sampling   & {\bf 30.6}  & {\bf 24.3} & {\bf 92.1}    \\
Uniform Scheduled Sampling   & 29.2  & 24.2 & 90.9    \\
\hline
Baseline ensemble of 10 & 30.7 & 25.1 & 95.7 \\
Scheduled Sampling ensemble of 5  & {\bf 32.3} & {\bf 25.4} & {\bf 98.7} \\
\hline
\end{tabular}
\end{table}

It's worth noting that we used our scheduled sampling approach to participate in the
2015 MSCOCO image captioning challenge~\cite{coco-challenge}
and ranked first in the final leaderboard.

\subsection{Constituency Parsing}
\label{sec:parsing}

Another less obvious connection with the {\em any-to-sequence} paradigm is constituency parsing. Recent work~\cite{vinyals:2014:arxiv} has proposed an interpretation of a parse tree as a sequence of linear ``operations'' that build up the tree. This linearization procedure allowed them to train a model that can map a sentence onto its parse tree without any modification to the any-to-sequence formulation.

The trained model has one layer of 512 LSTM cells and words are represented
by embedding vectors of size 512.
We used an attention mechanism similar to the one described
in~\cite{bahdanau:2015} which helps, when considering the next output token to
produce $y_t$, to focus on part of the input sequence only by applying a
softmax over the LSTM state vectors corresponding to the input sequence.
The input word dictionary contained around 90k
words, while the target dictionary contained 128 symbols used to describe
the tree.
We used an inverse sigmoid decay schedule for $\epsilon_i$ in the scheduled
sampling approach.

Parsing is quite different from image captioning as the function that one has to learn is almost deterministic. In contrast to an image having a large number of valid captions, most sentences have a unique parse tree (although some very difficult cases exist). Thus, the model operates almost deterministically, which can be seen by observing that the train and test perplexities are extremely low compared to image captioning (1.1 vs. 7).

This different operating regime makes for an interesting comparison, as one would not expect the baseline algorithm to make many mistakes. However, and as can be seen in Table~\ref{tab:parsing}, scheduled sampling has a positive effect which is additive to dropout. In this table we report the F1 score on the WSJ 22 development set~\cite{hovy-EtAl:2006:NAACL}. We should also emphasize that there are only 40k training instances, so overfitting contributes largely to the performance of our system. Whether the effect of sampling during training helps with regard to overfitting or the training/inference mismatch is unclear, but the result is positive and additive with dropout. Once again, a model trained by always sampling from itself
instead of using the groundtruth previous token as input yielded very bad
results, in fact so bad that the resulting trees were often not valid trees
(hence the ``-'' in the corresponding F1 metric).

\begin{table}[!ht]
\caption{F1 score (the higher the better) on the validation set of the parsing task.}\label{tab:parsing}
\centering
\begin{tabular}{|c|c|}
\hline
Approach & F1 \\
\hline
\hline
Baseline LSTM & 86.54  \\
Baseline LSTM with Dropout & 87.0  \\
Always Sampling &  -  \\
Scheduled Sampling & {\bf 88.08}    \\
Scheduled Sampling with Dropout & {\bf 88.68}    \\
\hline
\end{tabular}
\end{table}

\subsection{Speech Recognition}
\label{sec:speech}
For the speech recognition experiments, we used a slightly different setting from
the rest of the paper. Each training example is an input/output pair $(X,Y)$,
where $X$ is a sequence of $T$ input vectors $x_1,x_2,\cdots x_T$ and
$Y$ is a sequence of $T$ tokens $y_1,y_2,\cdots y_T$ so each $y_t$
is aligned with the corresponding $x_t$. Here, $x_t$
are the acoustic features represented by log Mel filter bank spectra at
frame $t$, and $y_t$ is the corresponding target. The targets used
were HMM-state labels generated from a GMM-HMM recipe, using the Kaldi
toolkit~\cite{Kaldi}
but could very well have been phoneme labels. This setting is different
from the other experiments in that the model we used is the following:

\begin{eqnarray}
\label{eq:next_step_speech}
\log P(Y|X;\theta) & = & \log P(y_1^T|x_1^T;\theta) \nonumber\\
       & = & \sum_{t=1}^T \log P(y_t|y_1^{t-1}, x_1^t;\theta) \nonumber\\
       & = & \sum_{t=1}^T \log P(y_t|h_t;\theta)
\end{eqnarray}

where $h_t$ is computed by a recurrent neural network as follows:
\begin{equation}
\label{eq:recurrence_speech}
h_t = \left\{\begin{array}{ll}
  f({\bf o}_h, S, x_1;\theta) & \mbox{if } t = 1, \\
  f(h_{t-1}, y_{t-1}, x_t;\theta) & \mbox{otherwise.}
  \end{array}\right.
\end{equation}
where ${\bf o}_h$ is a vector of 0's with same dimensionality as
$h_t$'s and $S$ is an extra token added to the dictionary to
represent the start of each sequence.

We generated data for these experiments using the
TIMIT\footnote{\url{https://catalog.ldc.upenn.edu/LDC93S1}.}
corpus and the KALDI toolkit as described in \cite{jaitlythesis}.
Standard configurations were used for the experiments - 40 dimensional log Mel filter
banks and their first and second order temporal derivatives were used as
inputs to each frame. 180 dimensional targets were generated for each time frame
using forced alignment to transcripts using a trained GMM-HMM system.
The training, validation and test sets have 3696, 400 and 192 sequences
respectively, and their average length was 304 frames.
The validation set was used to choose the best epoch in training, and the
model parameters from that epoch were used to evaluate the test set.

The trained models had two layers of 250 LSTM cells and
a softmax layer, for each of five configurations - a baseline configuration
where the ground truth was always fed to the model, a configuration
(Always Sampling)
where the model was only fed in its own predictions from the last time step,
and three scheduled sampling configurations (Scheduled Sampling 1-3), where
$\epsilon_i$ was ramped linearly from a maximum value to a minimum value over
ten epochs and then kept constant at the final value.  For each configuration,
we trained 3 models
and report average performance over them.  Training of each model was
done over frame targets from the GMM. The baseline configurations typically
reached the best validation accuracy after approximately 14 epochs whereas
the sampling models reached the best accuracy after approximately 9 epochs,
after which the validation accuracy decreased. This is probably because
the way we trained our models is not exact - it does not account for the
gradient of the sampling probabilities from which we sampled our targets.
Future effort at tackling this problem may further improve results.

Testing was done by finding the best sequence from beam search decoding (using a
beam size of 10 beams) and computing the error rate over the sequences. We
also report the next step error rate (where the model was fed in the ground
truth to predict the class of the next frame) for each of the models on the
validation set to summarize the performance of the models on the training
objective.  Table~\ref{tab:speech} shows a summary of the results

It can be seen that the baseline performs better next step prediction than the models
that sample the tokens for input. This is to be expected, since the former
has access to the groundtruth.
However, it can be seen that the models that
were trained with sampling perform better than the baseline during decoding.
It can also be seen that for this problem,
the ``Always Sampling'' model performs quite well. We hypothesize that this
has to do with the nature of the dataset. The HMM-aligned states have a lot of
correlation - the same state appears as the target for several frames, and
most of the states are constrained only to go to a subset of other states.
Next step prediction with groundtruth labels on this task ends up paying
disproportionate attention to the structure of the labels ($y_1^{t-1}$) and
not enough to the acoustics input ($x_1^t$). Thus it achieves very good
next step prediction error when the groundtruth sequence is fed in with the
acoustic information, but is not able to exploit the acoustic information
sufficiently when the groundtruth sequence is not fed in. For this model
the testing conditions are too far from the training condition for it to
make good predictions. The model that is only fed its own prediction (Always Sampling)
ends up exploiting all the information it can find in the acoustic signal, and
effectively ignores its own predictions to influence the next step prediction.
Thus at test time, it performs just as well as it does during training.
A model such as the attention model of~\cite{chorowski2014end} which predicts
phone sequences directly, instead of the highly redundant HMM state sequences,
would not suffer from this problem because it would need to exploit both the
acoustic signal and the language model sufficiently to make predictions.
Nevertheless, even in this setting, adding scheduled sampling
still helped to improve the decoding frame error rate.

Note that typically speech recognition experiments use HMMs to decode predictions
from neural networks in a hybrid model. Here we avoid using an HMM altogether
and hence we do not have the advantage of the smoothing that
results from the HMM architecture and the language models. Thus the results
are not directly comparable to the typical hybrid model results.

\begin{table}[!ht]
	\caption{Frame Error Rate (FER) on the speech recognition experiments.
		In next step prediction (reported on validation set) the ground truth
		is fed in to predict the next target like it is done during training.
		In decoding experiments (reported on test set), beam searching is done
		to find the best sequence. We report results on four different linear
		schedulings of sampling, where $\epsilon_i$ was ramped down linearly
		from $\epsilon_s$ to $\epsilon_e$. For the baseline, the model was
		only fed in the ground truth. See Section~\ref{sec:speech} for
		an analysis of the results.
}\label{tab:speech}
\centering
\begin{tabular}{|c|c|c|c|c|}
\hline
Approach  & $\epsilon_s$ & $\epsilon_e$ & Next Step FER & Decoding FER \\
\hline
\hline
Always Sampling & 0 & 0 & 34.6 &  35.8 \\
Scheduled Sampling 1 & 0.25 & 0 & 34.3 & {\bf 34.5} \\
Scheduled Sampling 2 & 0.5  & 0 & 34.1 & 35.0 \\
Scheduled Sampling 3 & 0.9  & 0.5 & 19.8 & 42.0 \\
Baseline LSTM & 1  & 1 & 15.0 & 46.0 \\
\hline
\end{tabular}
\end{table}

\section{Conclusion}
\label{sec:conclusion}
Using recurrent neural networks to predict sequences of tokens has many
useful applications like machine translation and image description. However,
the current approach to training them, predicting one token at a time,
conditioned on the state and the previous correct token, is different from how
we actually use them and thus is prone to the accumulation of errors along
the decision paths. In this paper, we proposed a {\em curriculum learning}
approach to slowly change the training objective from an easy task, where
the previous token is known, to a realistic one, where it is provided by the
model itself. Experiments on several sequence prediction tasks yield
performance improvements, while not incurring longer training
times. Future work includes back-propagating the errors through the
sampling decisions, as well as exploring better sampling strategies including
conditioning on some confidence measure from the model itself.


{\small
\bibliographystyle{unsrt}
\bibliography{biblio}

\begin{thebibliography}{10}

\bibitem{bengio:1994:ieee}
Y.~Bengio, P.~Simard, and P.~Frasconi.
\newblock Learning long term dependencies is hard.
\newblock {\em {IEEE} Transactions on Neural Networks}, 5(2):157--166, 1994.

\bibitem{hochreiter1997long}
S.~Hochreiter and J.~Schmidhuber.
\newblock Long short-term memory.
\newblock {\em Neural Computation}, 9(8), 1997.

\bibitem{sutskever:2014}
I.~Sutskever, O.~Vinyals, and Q.~Le.
\newblock Sequence to sequence learning with neural networks.
\newblock In {\em Advances in Neural Information Processing Systems, {NIPS}},
  2014.

\bibitem{vinyals:2014:arxiv}
O.~Vinyals, L.~Kaiser, T.~Koo, S.~Petrov, I.~Sutskever, and G.~Hinton.
\newblock Grammar as a foreign language.
\newblock In {\em arXiv:1412.7449}, 2014.

\bibitem{vinyals:2015}
O.~Vinyals, A.~Toshev, S.~Bengio, and D.~Erhan.
\newblock Show and tell: A neural image caption generator.
\newblock In {\em {IEEE} Conference on Computer Vision and Pattern Recognition,
  {CVPR}}, 2015.

\bibitem{donahue:2015}
J.~Donahue, L.~A. Hendricks, S.~Guadarrama, M.~Rohrbach, S.~Venugopalan,
  K.~Saenko, and T.~Darrell.
\newblock Long-term recurrent convolutional networks for visual recognition and
  description.
\newblock In {\em {IEEE} Conference on Computer Vision and Pattern Recognition,
  {CVPR}}, 2015.

\bibitem{bengio:2009:icml}
Y.~Bengio, J.~Louradour, R.~Collobert, and J.~Weston.
\newblock Curriculum learning.
\newblock In {\em Proceedings of the International Conference on Machine
  Learning, {ICML}}, 2009.

\bibitem{CRFS}
J.~D. Lafferty, A.~McCallum, and F.~C.~N. Pereira.
\newblock Conditional random fields: Probabilistic models for segmenting and
  labeling sequence data.
\newblock In {\em Proceedings of the Eighteenth International Conference on
  Machine Learning, {ICML}}, pages 282--289, San Francisco, CA, USA, 2001.
  Morgan Kaufmann Publishers Inc.

\bibitem{daume:2009}
H.~{Daum\'e III}, J.~Langford, and D.~Marcu.
\newblock Search-based structured prediction as classification.
\newblock {\em Machine Learning Journal}, 2009.

\bibitem{ross:2011}
S.~Ross, G.~J. Gordon, and J.~A. Bagnell.
\newblock A reduction of imitation learning and structured prediction to
  no-regret online learning.
\newblock In {\em Proceedings of the Workshop on Artificial Intelligence and
  Statistics, {AISTATS}}, 2011.

\bibitem{venkatraman:aaai:2015}
A.~Venkatraman, M.~Herbert, and J.~A. Bagnell.
\newblock Improving multi-step prediction of learned time series models.
\newblock In {\em Twenty-Ninth AAAI Conference on Artificial Intelligence,
  {AAAI}}, 2015.

\bibitem{collins:2004}
M.~Collins and B.~Roark.
\newblock Incremental parsing with the perceptron algorithm.
\newblock In {\em Proceedings of the Association for Computational Linguistics,
  {ACL}}, 2004.

\bibitem{goldberg:coling:2012}
Y.~Goldberg and J.~Nivre.
\newblock A dynamic oracle for arc-eager dependency parsing.
\newblock In {\em Proceedings of {COLING}}, 2012.

\bibitem{baidu}
J.~Mao, W.~Xu, Y.~Yang, J.~Wang, Z.~H. Huang, and A.~Yuille.
\newblock Deep captioning with multimodal recurrent neural networks (m-rnn).
\newblock In {\em International Conference on Learning Representations,
  {ICLR}}, 2015.

\bibitem{toronto}
R.~Kiros, R.~Salakhutdinov, and R.~Zemel.
\newblock Unifying visual-semantic embeddings with multimodal neural language
  models.
\newblock In {\em TACL}, 2015.

\bibitem{karpathy:2015:cvpr}
A.~Karpathy and F.-F. Li.
\newblock Deep visual-semantic alignments for generating image descriptions.
\newblock In {\em {IEEE} Conference on Computer Vision and Pattern Recognition,
  {CVPR}}, 2015.

\bibitem{MSR}
H.~Fang, S.~Gupta, F.~Iandola, R.~K. Srivastava, L.~Deng, P.~Dollar, J.~Gao,
  X.~He, M.~Mitchell, J.~C. Platt, C.~L. Zitnick, and G.~Zweig.
\newblock From captions to visual concepts and back.
\newblock In {\em {IEEE} Conference on Computer Vision and Pattern Recognition,
  {CVPR}}, 2015.

\bibitem{devicider}
R.~Vedantam, C.~L. Zitnick, and D.~Parikh.
\newblock {CIDEr}: Consensus-based image description evaluation.
\newblock In {\em {IEEE} Conference on Computer Vision and Pattern Recognition,
  {CVPR}}, 2015.

\bibitem{COCO}
T.-Y. Lin, M.~Maire, S.~Belongie, J.~Hays, P.~Perona, D.~Ramanan,
  P.~Doll{\'a}r, and C.~L. Zitnick.
\newblock Microsoft coco: Common objects in context.
\newblock {\em arXiv:1405.0312}, 2014.

\bibitem{ioffe:2015}
S.~Ioffe and C.~Szegedy.
\newblock Batch normalization: Accelerating deep network training by reducing
  internal covariate shift.
\newblock In {\em Proceedings of the International Conference on Machine
  Learning, {ICML}}, 2015.

\bibitem{coco-challenge}
Y.~Cui, M.~R. Ronchi, T.-Y. Lin, P.~Dollár, and L.~Zitnick.
\newblock Microsoft coco captioning challenge.
\newblock http://mscoco.org/dataset/\#captions-challenge2015, 2015.

\bibitem{bahdanau:2015}
D.~Bahdanau, K.~Cho, and Y.~Bengio.
\newblock Neural machine translation by jointly learning to align and
  translate.
\newblock In {\em International Conference on Learning Representations,
  {ICLR}}, 2015.

\bibitem{hovy-EtAl:2006:NAACL}
E.~Hovy, M.~Marcus, M.~Palmer, L.~Ramshaw, and R.~Weischedel.
\newblock Ontonotes: The 90\% solution.
\newblock In {\em Proceedings of the Human Language Technology Conference of
  the NAACL, Short Papers}, pages 57--60, New York City, USA, June 2006.
  Association for Computational Linguistics.

\bibitem{Kaldi}
D.~Povey, A.~Ghoshal, G.~Boulianne, L.~Burget, O.~Glembek, N.~Goel,
  M.~Hannemann, P.~Motlicek, Y.~Qian, P.~Schwarz, J.~Silovsky, G.~Stemmer, and
  K.~Vesely.
\newblock The kaldi speech recognition toolkit.
\newblock In {\em IEEE 2011 Workshop on Automatic Speech Recognition and
  Understanding}. IEEE Signal Processing Society, December 2011.
\newblock IEEE Catalog No.: CFP11SRW-USB.

\bibitem{jaitlythesis}
N.~Jaitly.
\newblock {\em Exploring Deep Learning Methods for discovering features in
  speech signals.}
\newblock PhD thesis, University of Toronto, 2014.

\bibitem{chorowski2014end}
Jan Chorowski, Dzmitry Bahdanau, Kyunghyun Cho, and Yoshua Bengio.
\newblock End-to-end continuous speech recognition using attention-based
  recurrent nn: First results.
\newblock {\em arXiv preprint arXiv:1412.1602}, 2014.

\end{thebibliography}
}

\end{document}